\documentclass[11pt,conference]{IEEEtran}
\IEEEoverridecommandlockouts

\usepackage{cite}
\usepackage{amsmath,amssymb,amsfonts}
\usepackage{graphicx}
\usepackage{textcomp}
\usepackage{xcolor}
\usepackage{booktabs}
\usepackage{array}
\usepackage{multirow}
\usepackage{url}
\usepackage{tikz}
\usetikzlibrary{positioning, arrows.meta, fit, backgrounds}
\def\BibTeX{{\rm B\kern-.05em{\sc i\kern-.025em b}\kern-.08em
    T\kern-.1667em\lower.7ex\hbox{E}\kern-.125emX}}

\begin{document}

\title{ERN-Net : Evolving Reason Node-Net for Document Binarization}

\author{
\IEEEauthorblockN{Hsin-Jui Pan}
\IEEEauthorblockA{
\textit{Dept. of Electrical Engineering} \\
\textit{Tamkang University} \\
412440314@o365.tku.edu.tw
}
\and
\IEEEauthorblockN{Sheng-Wei Chan}
\IEEEauthorblockA{
\textit{Dept. of Electrical Engineering} \\
\textit{Tamkang University} \\
412440330@o365.tku.edu.tw
}
\and
\IEEEauthorblockN{Jen-Shiun Chiang*}
\IEEEauthorblockA{
\textit{Dept. of Electrical Engineering} \\
\textit{Tamkang University} \\
jschiang@mail.tku.edu.tw
}
}

\maketitle

\begin{abstract}
This paper presents ERN-Net, an Evolving Reason Node-Net for efficient document image binarization. ERN-Net enhances degradation-sensitive regions, such as faint strokes, broken characters, and noisy backgrounds, through evolving reason nodes and multi-scale reasoning. We further compare ResNet-101, ConvNeXt-Tiny, and ConvNeXt-Base, and find that ConvNeXt-Tiny provides the best practical trade-off between accuracy and memory usage. In addition, DIBCO-based pretraining improves binarization performance without increasing model memory consumption, requiring only about 1.5 additional training hours. Experiments on DIBCO-style benchmarks show that ERN-Net is effective under low-data and low-memory settings.
\end{abstract}


\section{Introduction}

Document image binarization separates foreground text from degraded backgrounds and is widely used in optical character recognition, historical document restoration, and document understanding. Traditional thresholding methods, such as Otsu~\cite{otsu1979threshold} and Sauvola~\cite{sauvola2000adaptive}, are efficient but often fail on documents with uneven illumination, stains, bleed-through, and low-contrast handwriting. Deep learning methods improve binarization by learning pixel-wise foreground-background representations~\cite{tensmeyer2017document,vo2018binarization}. However, many models still process all spatial regions uniformly, although most binarization errors are concentrated around faint strokes, broken characters, text boundaries, and noisy backgrounds. Therefore, focusing feature reasoning on these difficult regions is important for both accuracy and efficiency. We propose ERN-Net, an Evolving Reason Node-Net for document image binarization. ERN-Net uses evolving reason nodes to enhance degradation-sensitive areas and adopts multi-scale reasoning for both local stroke details and broader background degradation. We also study backbone selection and DIBCO-based pretraining. The contributions are threefold: ERN-Net introduces reason-node-based binarization, identifies ConvNeXt-Tiny as an efficient backbone, and shows that DIBCO-based pretraining improves performance without increasing memory usage.

\section{Related Work}

\subsection{Document Image Binarization}

Document image binarization has been widely studied in document analysis. Otsu's method~\cite{otsu1979threshold} uses a global threshold selected by maximizing between-class variance. Sauvola's method~\cite{sauvola2000adaptive} computes adaptive local thresholds and is more suitable for nonuniform backgrounds. Howe~\cite{howe2011laplacian} formulated binarization as an energy minimization problem to improve text structure preservation. The DIBCO and H-DIBCO competitions are widely used benchmarks for degraded document binarization~\cite{gatos2009dibco,pratikakis2013icdar,pratikakis2017icdar,pratikakis2019icdar}. Common evaluation metrics include F-measure (FM), pseudo F-measure (p-FM), PSNR, and DRD.

\subsection{Deep Learning and Efficient Backbones}

Deep learning-based methods usually formulate binarization as a dense prediction task. Tensmeyer and Martinez~\cite{tensmeyer2017document} proposed a fully convolutional network for document image binarization. Later methods introduced hierarchical supervision, generative learning, and adaptive threshold-inspired networks~\cite{vo2018binarization,li2021sauvolanet}. Backbone design also affects binarization performance. ResNet~\cite{he2016deep} provides stable residual learning, while ConvNeXt~\cite{liu2022convnet} modernizes convolutional networks with stronger efficiency-performance trade-offs.

\section{Proposed Method}

\subsection{Overview}

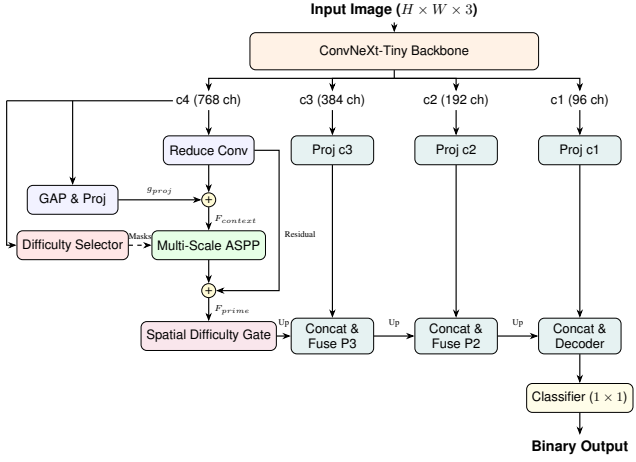
\begin{figure}[t]
\centering
\resizebox{0.95\linewidth}{!}{%
\begin{tikzpicture}[
    block/.style={
        draw,
        rectangle,
        rounded corners,
        minimum width=2.2cm,
        minimum height=0.7cm,
        text centered,
        align=center,
        fill=blue!5,
        font=\sffamily\footnotesize
    },
    fpn/.style={
        draw,
        rectangle,
        rounded corners,
        minimum width=2.0cm,
        minimum height=0.7cm,
        text centered,
        align=center,
        fill=teal!10,
        font=\sffamily\footnotesize
    },
    sum/.style={
        draw,
        circle,
        inner sep=1pt,
        fill=yellow!20,
        font=\footnotesize
    },
    arrow/.style={
        -{Stealth[scale=0.9]},
        thick
    },
    dashed_arrow/.style={
        -{Stealth[scale=0.9]},
        dashed,
        thick
    }
]

\node (input) at (1, 0) [font=\sffamily\small\bfseries] 
{Input Image ($H \times W \times 3$)};

\node[
    block,
    fill=orange!10,
    minimum height=0.9cm,
    minimum width=7cm
] (backbone) at (1, -1.0) 
{ConvNeXt-Tiny Backbone};

\draw[arrow] (input) -- (backbone);

\node[font=\sffamily\footnotesize, inner sep=2pt] (c4) at (-3.5, -2.2) {c4 (768 ch)};
\node[font=\sffamily\footnotesize, inner sep=2pt] (c3) at (-0.5, -2.2) {c3 (384 ch)};
\node[font=\sffamily\footnotesize, inner sep=2pt] (c2) at (2.5, -2.2) {c2 (192 ch)};
\node[font=\sffamily\footnotesize, inner sep=2pt] (c1) at (5.5, -2.2) {c1 (96 ch)};

\draw[thick] (backbone.south) -- ++(0, -0.2) coordinate (bus);
\draw[arrow] (bus) -| (c4.north);
\draw[arrow] (bus) -| (c3.north);
\draw[arrow] (bus) -| (c2.north);
\draw[arrow] (bus) -| (c1.north);

\node[block] (reduce) at (-3.5, -3.4) {Reduce Conv};
\node[block] (gap) at (-6.8, -4.6) {GAP \& Proj};
\node[block, fill=red!10] (selector) at (-6.8, -5.7) {Difficulty Selector};

\draw[arrow] (c4) -- (reduce);

\draw[thick] (c4.west) -- (-8.4, -2.2);
\draw[arrow] (-6.8, -2.2) -- (gap.north);
\draw[arrow] (-8.4, -2.2) |- (selector.west);

\node[sum] (sum1) at (-3.5, -4.6) {+};
\draw[arrow] (gap.east) -- (sum1.west)
    node[midway, above, font=\tiny] {$g_{proj}$};
\draw[arrow] (reduce.south) -- (sum1.north);

\node[block, fill=green!10] (aspp) at (-3.5, -5.7) {Multi-Scale ASPP};
\draw[arrow] (sum1.south) -- (aspp.north)
    node[midway, right, font=\tiny] {$F_{context}$};
\draw[dashed_arrow] (selector.east) -- (aspp.west)
    node[midway, above, font=\tiny] {Masks};

\node[sum] (sum2) at (-3.5, -6.8) {+};
\draw[arrow] (aspp.south) -- (sum2.north);

\draw[arrow] (reduce.east) -- ++(0.6,0) |- (sum2.east)
    node[pos=0.3, right, font=\tiny] {Residual};

\node[block, fill=purple!10] (gate) at (-3.5, -7.9) {Spatial Difficulty Gate};
\draw[arrow] (sum2.south) -- (gate.north)
    node[midway, right, font=\tiny] {$F_{prime}$};

\node[fpn] (proj3) at (-0.5, -3.4) {Proj c3};
\node[fpn] (proj2) at (2.5, -3.4) {Proj c2};
\node[fpn] (proj1) at (5.5, -3.4) {Proj c1};

\draw[arrow] (c3) -- (proj3);
\draw[arrow] (c2) -- (proj2);
\draw[arrow] (c1) -- (proj1);

\node[fpn, fill=teal!10] (fuse3) at (-0.5, -7.9) {Concat \& \\ Fuse P3};
\node[fpn, fill=teal!10] (fuse2) at (2.5, -7.9) {Concat \& \\ Fuse P2};
\node[fpn, fill=teal!10] (fuse1) at (5.5, -7.9) {Concat \& \\ Decoder};

\draw[arrow] (gate.east) -- (fuse3.west)
    node[midway, above=2pt, font=\tiny] {Up};
\draw[arrow] (proj3.south) -- (fuse3.north);

\draw[arrow] (fuse3.east) -- (fuse2.west)
    node[midway, above=2pt, font=\tiny] {Up};
\draw[arrow] (proj2.south) -- (fuse2.north);

\draw[arrow] (fuse2.east) -- (fuse1.west)
    node[midway, above=2pt, font=\tiny] {Up};
\draw[arrow] (proj1.south) -- (fuse1.north);

\node[block, fill=yellow!10] (classifier) at (5.5, -9.4) {Classifier ($1 \times 1$)};
\node[font=\sffamily\small\bfseries] (output) at (5.5, -10.6) {Binary Output};

\draw[arrow] (fuse1.south) -- (classifier.north);
\draw[arrow] (classifier.south) -- (output.north);

\end{tikzpicture}%
}
\caption{ERN-Net architecture. ConvNeXt-Tiny extracts hierarchical features; the evolving reason node selector, multi-scale reasoning, and spatial difficulty gate enhance degradation-sensitive regions before the progressive decoder fuses them with low-level features for binary prediction.}
\label{fig:architecture}
\end{figure}

ERN-Net follows an encoder-decoder architecture for binary document prediction. Given an input image $I \in \mathbb{R}^{H \times W \times 3}$, the network predicts a foreground probability map $P \in [0,1]^{H \times W}$. The final binary output is obtained by thresholding:

\begin{equation}
B(x,y)=
\begin{cases}
1, & P(x,y)\geq \tau,\\
0, & P(x,y)<\tau,
\end{cases}
\end{equation}

\noindent where $\tau$ is set to $0.5$. As shown in Fig.~\ref{fig:architecture}, a ConvNeXt-Tiny backbone extracts hierarchical features; the deepest feature is refined by the evolving reason node module, multi-scale reasoning, and a spatial difficulty gate, and is then fused with low-level features through a progressive decoder.

\subsection{Evolving Reason Node Module}

Let $F \in \mathbb{R}^{H' \times W' \times C}$ be the high-level feature map. ERN-Net predicts an importance map:

\begin{equation}
S = \sigma(\phi(F)),
\end{equation}

\noindent where $\phi(\cdot)$ is a lightweight selector. The map $S$ highlights difficult regions such as faint strokes, boundaries, and degraded backgrounds. A quota-based reason-node mask is generated as:

\begin{equation}
M = \mathcal{R}(S),
\end{equation}

\noindent where $\mathcal{R}(\cdot)$ denotes Top-K reason-node selection. A straight-through estimator is used so that the forward pass uses a binary mask while gradients still flow through the selector. The enhanced feature is:

\begin{equation}
F_{ern}=F\odot M+F.
\end{equation}

This residual design strengthens important regions without discarding the original feature representation.

\subsection{Multi-Scale Reasoning and Spatial Gate}

ERN-Net uses multi-scale branches to capture both local strokes and broader degradation patterns:

\begin{equation}
F_i=\mathcal{C}_i(F_{ern}),
\end{equation}

\noindent where $\mathcal{C}_i(\cdot)$ denotes a convolution branch. The branches include a $1\times1$ convolution and dilated $3\times3$ convolutions. Their outputs are fused as:

\begin{equation}
F_{agg}=F_{ern}+\sum_i F_i.
\end{equation}

A spatial difficulty gate further reweights the feature:

\begin{equation}
F_{gate}=F_{agg}\odot G(F_{agg})+F_{agg}.
\end{equation}

Finally, a progressive decoder fuses multi-level features and predicts the binary map:

\begin{equation}
P=\sigma(\mathcal{D}(F_{gate},F_{low})).
\end{equation}

\begin{table}[!t]
\caption{Reasoning branches in ERN-Net.}
\label{tab:multiscale_roles}
\centering
\scriptsize
\setlength{\tabcolsep}{2.8pt}
\begin{tabular}{ccc}
\toprule
\textbf{Branch} & \textbf{Role} & \textbf{Node Quota} \\
\midrule
$1\times1$ & Pixel-level refinement & 2000 \\
$3\times3$, dilated & Local degradation & 1500 \\
$3\times3$, dilated & Broken strokes & 800 \\
$3\times3$, dilated & Background context & 400 \\
\bottomrule
\end{tabular}
\end{table}

\subsection{Training Objective}

The training objective combines hard-pixel binarization loss and boundary supervision:

\begin{equation}
\mathcal{L}=\mathcal{L}_{OHEM}+\lambda_b\mathcal{L}_{bd},
\end{equation}

\noindent where $\mathcal{L}_{OHEM}$ is online hard example mining binary cross-entropy loss, and $\mathcal{L}_{bd}$ encourages the selector to focus on text boundaries. $\lambda_b$ is set to $1.5$.

\subsection{DIBCO-Based Pretraining}
\begin{figure}[!t]
\centering
\includegraphics[width=0.48\linewidth]{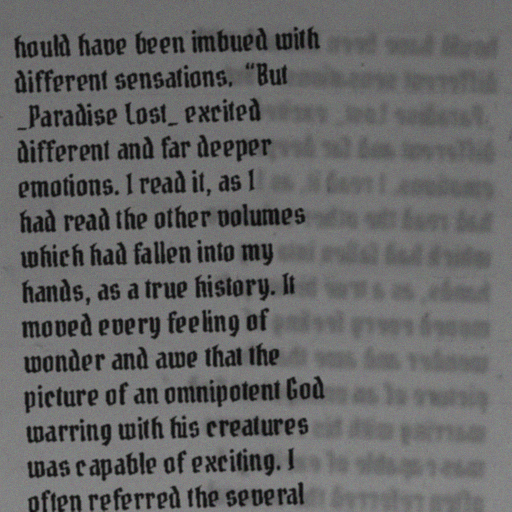}
\hfill
\includegraphics[width=0.48\linewidth]{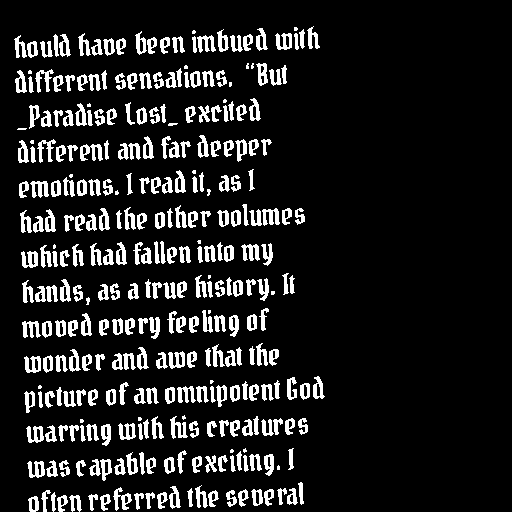}
\caption{Examples of DIBCO-based pretraining samples. The synthetic data provide task-specific initialization without changing the ERN-Net architecture or increasing memory usage.}
\label{fig:pretrain_samples}
\end{figure}

We construct a lightweight DIBCO-based pretraining stage using DIBCO 2009, H-DIBCO 2010, H-DIBCO 2012, and additional DIBCO-style training samples. Synthetic degraded samples are generated by combining clean text, estimated paper backgrounds, and degradation effects such as bleed-through, blur, illumination change, and noise, as illustrated in Fig.~\ref{fig:pretrain_samples}. This pretraining is used only for initialization before fine-tuning. Therefore, it does not introduce extra modules or increase memory usage during training and inference. In our setting, the pretraining stage requires only about 1.5 additional hours on an RTX 3060 Ti.

\section{Experiments}

\subsection{Experimental Setup}

We train ERN-Net using DIBCO 2009, H-DIBCO 2010, H-DIBCO 2012, and additional DIBCO-style training samples, under a single-GPU setting adapted to binary document prediction. The goal is to evaluate whether limited DIBCO data can provide effective pretraining and fine-tuning. All experiments are conducted on an NVIDIA RTX 3060 Ti GPU. The input crop size is $512\times512$, the batch size is 4, and gradient accumulation is set to 4, resulting in an effective batch size of 16. We train the model for 150 epochs using AdamW. The learning rate is $1\times10^{-5}$ for the backbone and $1\times10^{-4}$ for the head, with a weight decay of $1\times10^{-4}$. Automatic mixed precision is used to reduce memory consumption. Data augmentation includes random crop, horizontal and vertical flip, random rotation, brightness and contrast adjustment, color jitter, Gaussian noise, blur, and coarse dropout. During early training, label dilation is applied to stabilize thin stroke learning and is removed afterward. The training set is split into 80\% training and 20\% validation with a fixed random seed. The best checkpoint is selected according to validation F-measure.

\subsection{Quantitative Results}

Table~\ref{tab:dibco_results} reports the per-dataset binarization results of ERN-Net across seven DIBCO benchmarks. Higher FM, p-FM, and PSNR are better, while lower DRD is better.

\begin{table}[!t]
\caption{Per-dataset DIBCO evaluation of ERN-Net (ConvNeXt-Tiny backbone, with DIBCO-based pretraining).}
\label{tab:dibco_results}
\centering
\scriptsize
\setlength{\tabcolsep}{5.0pt}
\begin{tabular}{lcccc}
\toprule
\textbf{Dataset} & \textbf{FM (\%)} & \textbf{p-FM (\%)} & \textbf{PSNR} & \textbf{DRD} \\
\midrule
DIBCO 2011   & 93.55 & 95.76 & 19.60 & 1.5388 \\
DIBCO 2013   & 94.05 & 95.83 & 21.03 & 1.4447 \\
H-DIBCO 2014 & 95.85 & 97.24 & 21.98 & 0.8538 \\
H-DIBCO 2016 & 87.88 & 92.50 & 16.86 & 3.4900 \\
DIBCO 2017   & 91.61 & 93.49 & 17.69 & 2.5180 \\
H-DIBCO 2018 & 88.49 & 92.42 & 17.80 & 3.0945 \\
DIBCO 2019   & 74.00 & 74.50 & 13.41 & 8.0067 \\
\midrule
\textbf{Average} & \textbf{89.35} & \textbf{91.68} & \textbf{18.34} & \textbf{2.9924} \\
\bottomrule
\end{tabular}
\end{table}

\begin{figure*}[!t]
\centering
\includegraphics[width=0.48\linewidth]{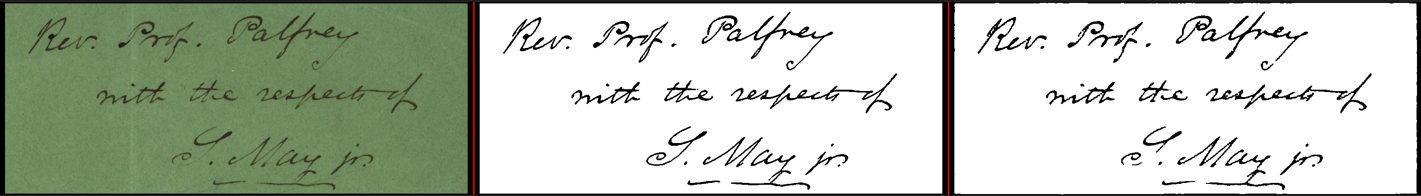}
\hfill
\includegraphics[width=0.48\linewidth]{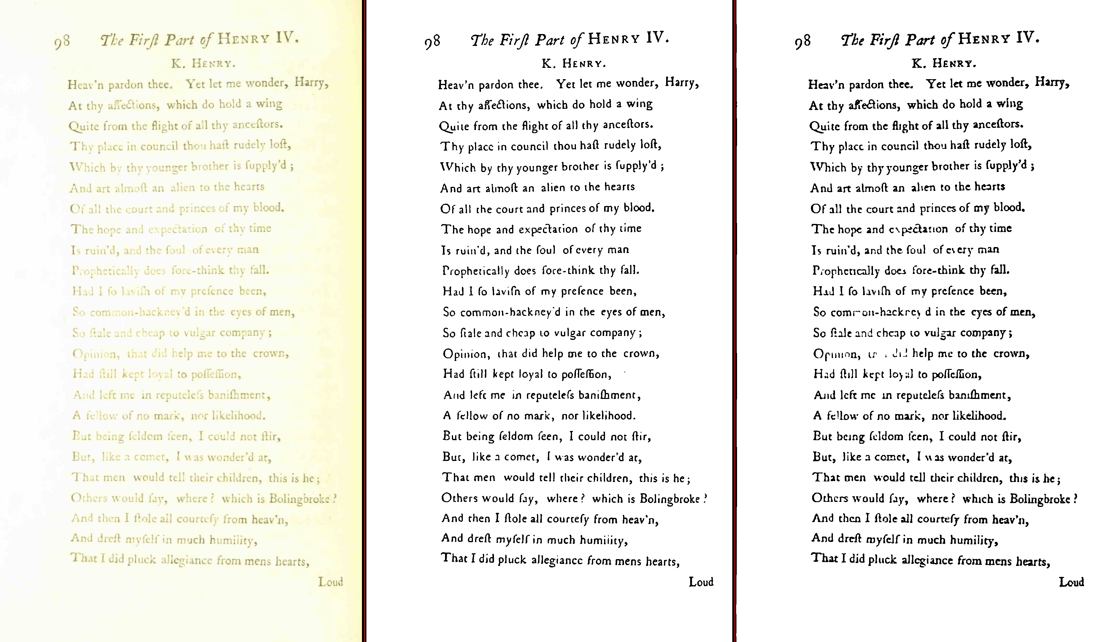}
\caption{Qualitative binarization results on degraded DIBCO images. Left: input image; middle: ground truth; right: ERN-Net prediction.}
\label{fig:qualitative_results}
\end{figure*}

As shown in Fig.~\ref{fig:qualitative_results}, ERN-Net preserves weak text strokes while suppressing noisy background regions on heavily degraded documents.

\subsection{Backbone and Pretraining Analysis}

Table~\ref{tab:backbone} compares ResNet-101, ConvNeXt-Tiny, and ConvNeXt-Base under the same pipeline. ConvNeXt-Base obtains slightly higher scores, but it requires more memory. ConvNeXt-Tiny achieves a better practical trade-off and is therefore selected as the default backbone.

\begin{table}[!t]
\caption{Backbone comparison on DIBCO-style binarization (average over all test sets).}
\label{tab:backbone}
\centering
\scriptsize
\setlength{\tabcolsep}{2.5pt}
\begin{tabular}{lccccc}
\toprule
\textbf{Backbone} & \textbf{Memory} & \textbf{FM} & \textbf{p-FM} & \textbf{PSNR} & \textbf{DRD} \\
\midrule
ResNet-101    & 12GB & 81.36 & 82.57 & 15.92 & 7.3065 \\
ConvNeXt-Tiny & \textbf{8GB} & 89.35 & 91.68 & 18.34 & 2.9924 \\
ConvNeXt-Base & 12GB & \textbf{90.01} & \textbf{91.96} & \textbf{18.75} & \textbf{2.5914} \\
\bottomrule
\end{tabular}
\end{table}

Table~\ref{tab:pretrain} shows the effect of DIBCO-based pretraining. The pretrained model improves all metrics while keeping the same architecture and memory usage. The only extra cost is the pretraining stage, which takes about 1.5 additional hours on an RTX 3060 Ti.

\begin{table}[!t]
\caption{Effect of DIBCO-based pretraining (ConvNeXt-Tiny backbone, average over all test sets).}
\label{tab:pretrain}
\centering
\scriptsize
\setlength{\tabcolsep}{2.6pt}
\begin{tabular}{lccccc}
\toprule
\textbf{Pretrain} & \textbf{Memory} & \textbf{FM} & \textbf{p-FM} & \textbf{PSNR} & \textbf{DRD} \\
\midrule
No  & 8GB & 87.12 & 87.94 & 17.23 & 3.7826 \\
Yes & 8GB & \textbf{89.35} & \textbf{91.68} & \textbf{18.34} & \textbf{2.9924} \\
\bottomrule
\end{tabular}
\end{table}

\section{Conclusion}

This paper presented ERN-Net for efficient document image binarization. ERN-Net enhances degradation-sensitive regions using evolving reason nodes and multi-scale reasoning. Experiments show that ConvNeXt-Tiny provides a favorable balance between accuracy and memory usage. DIBCO-based pretraining further improves performance without increasing model memory, requiring only about 1.5 additional training hours. Future work will evaluate ERN-Net on more diverse historical document datasets.

\section*{Acknowledgment}

\noindent This research work is partially supported by the National Science and Technology Council, Taiwan, under grant number 114-2221-E-032-011.

\end{document}